%% file: camera_ready.tex
\documentclass[10pt,twocolumn,letterpaper]{article}

\usepackage[pagenumbers]{cvpr} 

\input{preamble}

\definecolor{cvprblue}{rgb}{0.21,0.49,0.74}
\usepackage[pagebackref,breaklinks,colorlinks,allcolors=cvprblue]{hyperref}

\def\paperID{} 
\def\confName{CVPR}
\def\confYear{2026}

\title{Denoising as Path Planning: Training-Free Acceleration of Diffusion Models with DPCache}

\author{
 Bowen Cui\thanks{Equal contribution. $^\dagger$Corresponding authors.} \quad
 Yuanbin Wang$^*$ \quad
 Huajiang Xu$^\dagger$ \quad
 Biaolong Chen \\
 Aixi Zhang$^\dagger$ \quad
 Hao Jiang \quad
 Zhengzheng Jin \quad
 Xu Liu \quad
 Pipei Huang \\
 Alibaba Group
}

\begin{document}
\maketitle
\input{paper_sec/0_abstract}  
\input{paper_sec/1_intro}
\input{paper_sec/2_related}
\input{paper_sec/3_method}

\input{paper_sec/4_exp}

\input{paper_sec/5_conclusion}
{
    \small
    \bibliographystyle{ieeenat_fullname}
    \bibliography{ref}
}

\clearpage
\input{paper_sec/appendix}

\end{document}

%% file: preamble.tex
\usepackage{booktabs}      
\usepackage{multirow}      
\usepackage{amsmath}       
\usepackage{array}         
\usepackage[table]{xcolor}
\usepackage{pifont}
\usepackage{adjustbox}
\usepackage{makecell}
\usepackage{algorithm}
\usepackage{algpseudocode}
\usepackage{graphicx}
\usepackage{subcaption}
\definecolor{mygray}{gray}{.9}

\newcommand{\firstsentence}[1]{\noindent\textbf{#1}}



%% file: paper_sec/0_abstract.tex
\begin{abstract}
Diffusion models have demonstrated remarkable success in image and video generation, yet their practical deployment remains hindered by the substantial computational overhead of multi-step iterative sampling.
Among acceleration strategies, caching-based methods offer a training-free and effective solution by reusing or predicting features across timesteps. 
However, existing approaches rely on fixed or locally adaptive schedules without considering the global structure of the denoising trajectory, often leading to error accumulation and visual artifacts. 
To overcome this limitation, we propose DPCache, a novel training-free acceleration framework that formulates diffusion sampling acceleration as a global path planning problem. 
DPCache constructs a Path-Aware Cost Tensor from a small calibration set to quantify the path-dependent error of skipping timesteps conditioned on the preceding key timestep.
Leveraging this tensor, DPCache employs dynamic programming to select an optimal sequence of key timesteps that minimizes the total path cost while preserving trajectory fidelity.
During inference, the model performs full computations only at these key timesteps, while intermediate outputs are efficiently predicted using cached features. 
Extensive experiments on DiT, FLUX, and HunyuanVideo demonstrate that DPCache achieves strong acceleration with minimal quality loss, outperforming prior acceleration methods by $+$0.031 ImageReward at 4.87$\times$ speedup and even surpassing the full-step baseline by $+$0.028 ImageReward at 3.54$\times$ speedup on FLUX, validating the effectiveness of our path-aware global scheduling framework.
Code is available at \url{https://github.com/argsss/DPCache}.
\end{abstract}

%% file: paper_sec/1_intro.tex
\section{Introduction}
\label{sec:intro}

In recent years, diffusion models~\cite{ho2020denoising, rombach2022high, flux2024, wan2025, kong2024hunyuanvideo} have achieved remarkable success in image and video generation, leading to widespread applications in domains such as robotics, intelligent content creation, and virtual reality.
However, diffusion models, particularly diffusion transformers (DiTs)~\cite{peebles2023scalable}, are often characterized by their large-scale architectures and reliance on multi-step iterative sampling during inference, resulting in substantial computational overhead and prolonged latency, which severely hinder their practical deployment in real-world scenarios. 
As a result, accelerating the sampling process of diffusion models has emerged as a critical research direction to enable their broader adoption.

\begin{figure}[t]
    \centering
    \begin{subfigure}[b]{0.48\linewidth}
        \centering
        \includegraphics[width=\linewidth]{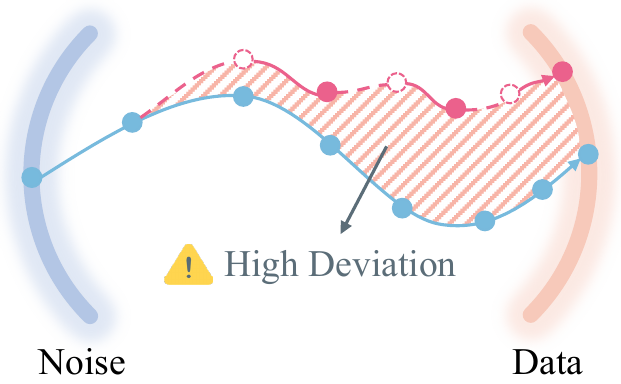}
        \caption{Fixed schedule}
        \label{fig:intro_a}
    \end{subfigure}
    \hfill
    \begin{subfigure}[b]{0.48\linewidth}
        \centering
        \includegraphics[width=\linewidth]{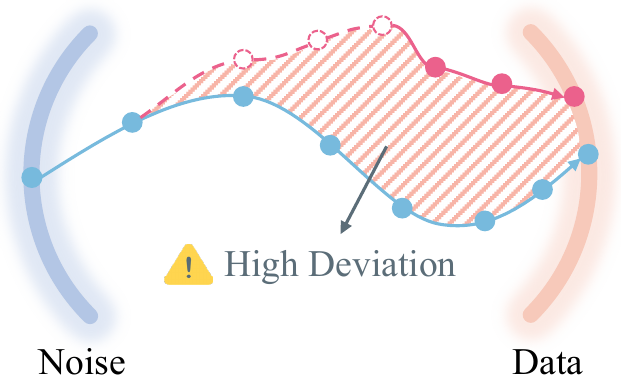}
        \caption{Locally adaptive schedule}
        \label{fig:intro_b}
    \end{subfigure}
    \vspace{0.5em}

    \begin{subfigure}[b]{0.7428\linewidth}
        \centering
        \includegraphics[width=\linewidth]{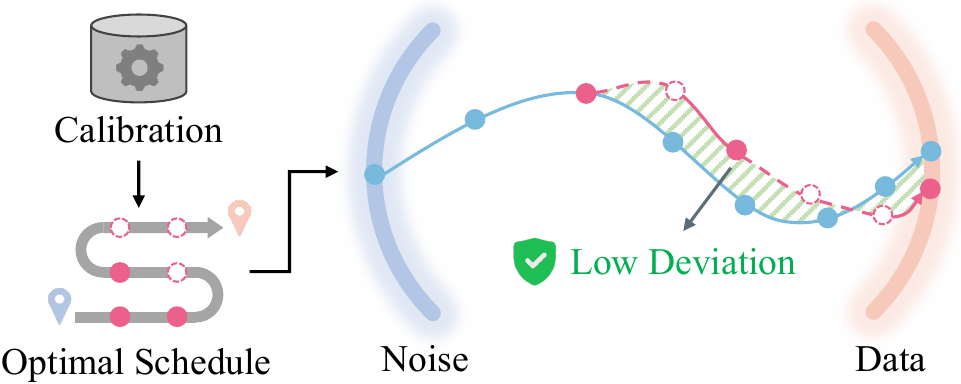}
        \caption{DPCache}
        \label{fig:intro_c}
    \end{subfigure}
    \hfill
    \begin{subfigure}[b]{0.2172\linewidth}
        \centering
        \includegraphics[width=\linewidth]{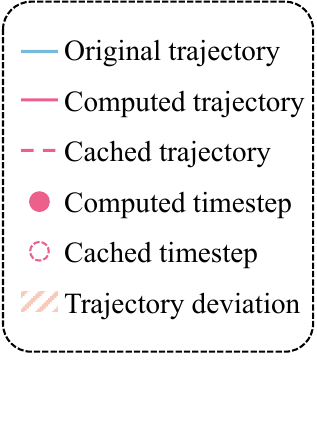}
    \end{subfigure}
    \caption{(a) Fixed schedule is inflexible and unable to identify critical timesteps, resulting in large deviation from the true trajectory.
    (b) Locally adaptive schedule makes greedy, short-sighted decisions that often skip essential timesteps, leading to irreversible deviation.
    (c) DPCache identifies a globally optimal sequence of key timesteps through calibration and achieves low cumulative trajectory deviation. }
    \vspace{-0.1cm}
    \label{fig:intro}
\end{figure}

Existing acceleration approaches for diffusion models can be broadly categorized into two paradigms: (i) reducing the number of sampling steps~\cite{song2020denoising, lu2022dpm, salimans2022progressive, watson2021learning, chen2024trajectory} and (ii) optimizing per-step computation~\cite{castells2024ld, feng2024relational, bolya2023token, ma2024deepcache, liu2025reusing}.
Among these, caching-based methods~\cite{ma2024deepcache, zou2024accelerating, liu2025timestep, liu2025speca, zhao2024real, liu2025reusing, feng2025hicache} have gained prominence due to their training-free nature and practical effectiveness.
DeepCache~\cite{ma2024deepcache} first observes that high-level features in diffusion models exhibit significant similarity across adjacent timesteps, allowing the reuse of previously computed features without significant degradation in the generation quality.
Subsequent works refine this paradigm by selectively caching tokens~\cite{zou2024accelerating}, varying update frequencies across modules~\cite{zhao2024real}, or dynamically deciding whether to reuse cached features at each timestep~\cite{liu2025timestep}. 
TaylorSeer~\cite{liu2025reusing} advances beyond simple reuse by predicting features at future timesteps with Taylor series expansion, capitalizing on the smooth evolution of features along the denoising trajectory.
However, prior caching-based methods rely on either fixed or locally adaptive sampling schedules, which neglect the global structure of the denoising trajectory.
As illustrated in Figure~\ref{fig:intro_a}, fixed-schedule approaches impose uniform skipping patterns irrespective of local feature dynamics, resulting in significant deviations in critical transition regions.
Figure~\ref{fig:intro_b} shows that locally adaptive strategies often make short-sighted decisions that skip critical timesteps, resulting in irreversible drift and error accumulation across the denoising process.
                   


To address these limitations, we argue that optimal acceleration requires a global perspective that accounts for the entire trajectory and plans the sampling steps holistically.
To this end, we propose \textbf{DPCache}, a novel training-free acceleration framework for diffusion models that formulates sampling acceleration as a global path planning problem.
As illustrated in Figure~\ref{fig:intro_c}, our approach begins by running the full denoising process on a small set of calibration samples to construct a \textbf{P}ath-\textbf{A}ware \textbf{C}ost \textbf{T}ensor (PACT).
The PACT captures the error of skipping between timesteps while accounting for the influence of the preceding key step, thereby reflecting the path-dependent nature of feature prediction.
Leveraging the PACT, we employ dynamic programming to select a sequence of key timesteps that minimizes the total path cost, resulting in a sparse yet high‑fidelity sampling schedule that preserves the structural integrity of the denoising process.
During inference, the denoising model performs full computation at the selected key timesteps, while efficiently predicting outputs for the remaining timesteps using cached features with off-the-shelf prediction methods~\cite{feng2025hicache, liu2025reusing}.
Since the prediction steps introduce negligible computational overhead, DPCache achieves significant acceleration.
Moreover, by selecting key timesteps through global path planning, DPCache produces a sampling trajectory that aligns more closely with the original denoising trajectory than prior methods, thereby maintaining generation quality comparable to the full-step baseline. 

We conduct extensive experiments on state-of-the-art diffusion models, including FLUX.1-dev~\cite{flux2024}, HunyuanVideo~\cite{kong2024hunyuanvideo} and DiT-XL~\cite{peebles2023scalable}.
DPCache consistently outperforms existing caching-based acceleration methods in both efficiency and generation quality, and achieves performance comparable to or even better than that of the full-step baseline.
For instance, on FLUX, DPCache yields a $+$0.031 higher ImageReward than the strongest prior method at a 4.87$\times$ speedup, and further improves upon the full-step baseline by $+$0.028 ImageReward at a 3.54$\times$ acceleration, demonstrating the effectiveness of our globally optimal sampling schedule.

In summary, our contributions are as follows:
\begin{itemize}
    \item We propose DPCache, a novel diffusion acceleration framework that formulates sampling acceleration as a global path planning problem, selecting an optimal sequence of key timesteps to minimize deviation from the full-step denoising trajectory.
    \item We propose a Path-Aware Cost Tensor that quantifies the path-dependent error of skipping intermediate timesteps along the denoising trajectory, and employ dynamic programming to efficiently identify the optimal sampling schedule for a fixed number of sampling steps. 
    \item DPCache achieves up to 4.87$\times$ acceleration across image and video generation models including DiT, FLUX, and HunyuanVideo, consistently outperforming prior caching-based methods in both efficiency and generation quality, establishing a new state of the art in training-free diffusion sampling acceleration.
\end{itemize}

%% file: paper_sec/2_related.tex
\section{Related Works}
\label{sec: related}


\subsection{Sampling Step Reduction}

\textbf{Dynamics redesign} accelerates diffusion by redefining the sampling trajectory or numerical integration scheme. 
DDIM~\cite{song2020denoising} introduces a deterministic sampling trajectory, allowing high-quality generation with fewer steps.
The DPM-Solver family~\cite{lu2022dpm, lu2025dpm, zheng2023dpm} employs high-order numerical solvers for ordinary differential equations (ODEs), significantly reducing the number of sampling steps.
Rectified Flow~\cite{liu2022flow} learns a straight transport path from noise to data by modeling instantaneous velocity,
while MeanFlow~\cite{geng2025mean} replaces this with average velocity to promote stability and improve fidelity in generation.

\textbf{Step distillation} reduces sampling steps by distilling a full-step teacher model into a few-step student.
Progressive distillation~\cite{salimans2022progressive} establishes this paradigm by iteratively compressing step counts through trajectory-level supervision.
Subsequent methods refine the distillation objective: Distribution Matching Distillation (DMD)~\cite{chadebec2025flash, yin2024improved, yin2024one} aligns the student and the teacher at the marginal distribution level, while adversarial diffusion distillation~\cite{lin2024sdxl, sauer2024adversarial, xu2024ufogen, chadebec2025flash} incorporates GAN-based discriminators~\cite{goodfellow2020generative} to improve perceptual fidelity. 
In parallel, consistency models~\cite{song2023consistency, wang2024phased, luo2023latent, mao2025osv} eliminate the need for a teacher altogether, instead enforcing the self-consistency of network predictions across timesteps.
However, these methods require additional training and are thus computationally expensive.

\subsection{Per-Step Computation Optimization}

\textbf{Denoising model compression} reduces computational overhead by compressing the model through parameter reduction strategies.
Pruning methods~\cite{zhang2024effortless, zhang2024laptop, castells2024ld, fang2023structural} sparsify weights or attention structures to reduce parameter count.
Quantization~\cite{li2023q, shang2023post, wu2024ptq4dit, kim2025ditto} enables low-bit inference by approximating full-precision parameters with compact numeric representations.
Some approaches leverage knowledge distillation~\cite{zhang2024accelerating, feng2024relational, li2023snapfusion} to transfer the knowledge of a larger teacher model to a lightweight student model. 
In parallel, token merging methods~\cite{bolya2023token, fang2025attend} dynamically fuse semantically similar tokens to shorten the sequence length and alleviate the computational cost of attention.  
Nevertheless, these methods typically introduce irreversible degradation in representational capacity, often necessitating nontrivial fine-tuning or retraining to recover generation quality.

\textbf{Feature caching} accelerates diffusion sampling by reusing previously computed intermediate features across timesteps, thereby avoiding redundant forward passes.
DeepCache~\cite{ma2024deepcache} first demonstrates that high-level feature maps in U-Net-based diffusion models exhibit strong inter-step similarity and can be safely reused with minimal quality degradation. 
This insight has since been extended to Diffusion Transformers~\cite{chen2024delta, selvaraju2024fora}.
To enhance robustness, adaptive strategies such as AdaCache~\cite{kahatapitiya2024adaptive} and TeaCache~\cite{liu2025timestep} further refine reuse decisions based on feature distance or predicted deviation.
Parallel efforts focuses on mitigating errors introduced by caching.
For instance, GOC~\cite{qiu2025accelerating} corrects feature drift via gradient-based propagation, while OmniCache~\cite{chu2025omnicache} employs stage-aware filtering to suppress cache-induced noise.
Recently, TaylorSeer~\cite{liu2025reusing} proposes a cache-then-forecast paradigm that predicts future features via Taylor expansion.
HiCache~\cite{feng2025hicache} improves the prediction accuracy with Hermite polynomials.
SpeCa~\cite{liu2025speca} employs a lightweight verifier network to dynamically decide whether to accept the predicted features.
Despite these advances, existing caching-based methods typically rely on fixed or locally adaptive schedules. 
Our proposed DPCache reframes sampling acceleration as a global path planning problem, explicitly optimizing the sequence of caching and computation decisions to minimize overall error accumulation.

%% file: paper_sec/3_method.tex
\section{Method}
\label{sec: method}

\subsection{Preliminary}
\firstsentence{Diffusion models.}
Diffusion models are a class of generative models that synthesize realistic samples through a forward diffusion process and a reverse denoising process.
The forward diffusion process is a Markov chain that progressively adds Gaussian noise to a clean data point $\boldsymbol{x}_0$ over $T$ timesteps according to a predefined noise schedule $\{\beta_t\}_{t=1}^{T}$:
\begin{equation}
    q(\boldsymbol{x}_t|\boldsymbol{x}_{t-1}) = \mathcal{N}(\boldsymbol{x}_t;\sqrt{1-\beta_t}\boldsymbol{x_{t-1}},\beta_t\boldsymbol{I})).
\end{equation}
After $T$ timesteps, the latent $\boldsymbol{x}_T$ is approximately distributed as standard Gaussian noise.
The reverse process aims to recover data by iteratively denoising from $\boldsymbol{x}_T\sim\mathcal{N}(0,\boldsymbol{I})$ back to $\boldsymbol{x}_0$.
It is modeled as a learned Markov chain with Gaussian transitions:
\begin{equation}
    p_{\theta}(\boldsymbol{x}_{t-1}|\boldsymbol{x}_{t}) = \mathcal{N}(\boldsymbol{x}_{t-1}; \boldsymbol{\mu}_{\theta}(\boldsymbol{x}_t, t), \boldsymbol{\Sigma}_{\theta}(\boldsymbol{x}_t, t)),
\end{equation}
where $\boldsymbol{\mu}_{\theta}$ and $\boldsymbol{\Sigma}_{\theta}$ are trained to approximate the true reverse dynamics.
As sampling involves $T$ sequential timesteps, fast and accurate sampling remains a fundamental requirement for diffusion models.


\firstsentence{Feature caching in diffusion models.}
Existing feature caching methods generally adopt a periodic temporal caching strategy with interval $\mathcal{N}$.
At a selected timestep $t$, the model performs a full forward pass and caches the output of each layer $l$ as $\boldsymbol{h}_t^l=\mathcal{F}^l(\boldsymbol{x}_t)$.
The cached feature stack $\mathcal{H} = \{\boldsymbol{h}_t^l\}_{l=1}^L$ is then reused for the next $\mathcal{N}-1$ timesteps without further computation: 
\begin{equation}
    \mathcal{F}^l(\boldsymbol{x}_{t-k}) = \boldsymbol{h}_t^l, k=1, 2, ..., \mathcal{N}-1.
\end{equation}
TaylorSeer~\cite{liu2025reusing} addresses the error accumulation inherent in naïve reuse by introducing Taylor-series-based predictor.
It maintains a history of finite differences up to order $m$, forming an enriched cache:
\begin{equation}
    \mathcal{H} = \{\boldsymbol{h}_t^l, \Delta\boldsymbol{h}_t^l, ..., \Delta^m\boldsymbol{h}_t^l\}_{l=1}^L.
\end{equation}
TaylorSeer uses the cached differences to approximate features at timestep $t-k$ via a truncated Taylor expansion:
\begin{equation}
\label{eq:taylor_pred}
    \mathcal{F}_{pred, m}^l(\boldsymbol{x}_{t-k}) = \boldsymbol{h}_t^l + \sum_{i=1}^{m}\frac{\Delta^i\boldsymbol{h}_t^l}{i!\cdot \mathcal{N}^i}(-k)^i.
\end{equation}
Our method builds on this predictive paradigm but uses a globally optimized caching strategy, instead of fixed or locally adaptive schedules.

\begin{figure*}
    \centering
    \includegraphics[width=\linewidth]{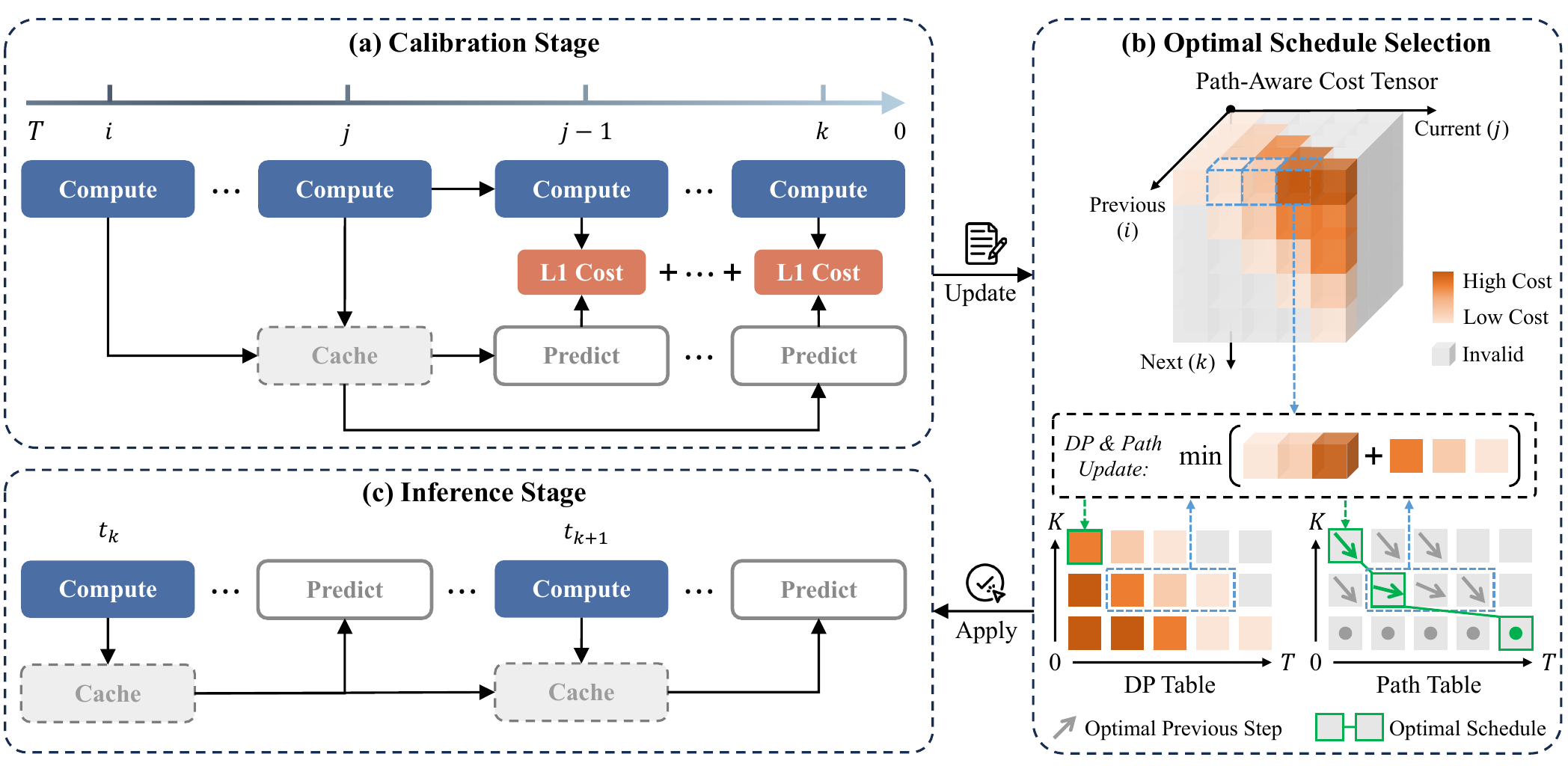}
    \caption{Overview of DPCache.
    (a) During the calibration stage, the full $T$-step denoising process is executed to construct a 3D Path-Aware Cost Tensor (PACT), which quantifies the cumulative error of skipping intermediate timesteps conditioned on the preceding key step.
    (b) An optimal $K$-step sampling schedule ($K < T$) is selected via dynamic programming, leveraging the PACT to maintain a DP table to store minimal cumulative costs and a path table to enable backtracking.
    (c) During inference, the model computes and caches features only at the preselected key timesteps, while features at other timesteps are efficiently predicted.}
    \label{fig:method}
\end{figure*}

\subsection{Overview}
\label{sec:overview}
Existing caching-based acceleration methods typically optimize the sampling schedule without accounting for the global structure of the denoising trajectory. 
To address this limitation, we propose \textbf{DPCache}, a framework that formulates diffusion sampling acceleration as a global path planning problem over the denoising trajectory.
As observed in prior work~\cite{chu2025omnicache}, the shape of the denoising trajectory is largely independent of the generated content and depends primarily on the diffusion model itself.
This enables us to calibrate an optimal sampling schedule on a small set of representative samples and apply it to arbitrary inputs.

As illustrated in Figure~\ref{fig:method}, DPCache operates as follows:
\begin{itemize}
    \item \textbf{Calibration stage}: We run the diffusion model on a small calibration set $\mathcal{S}_{\rm calib}$ using the full $T$-step denoising process.
    The cached features are used to construct a path-aware cost tensor that quantifies the error of skipping from one timestep to another, conditioned on the preceding key step.
    \item \textbf{Optimal schedule selection}: Given a target number of sampling steps $K<T$, we use dynamic programming to select an optimal timestep sequence $\mathcal{T}$ that minimizes the overall cost.
    \item \textbf{Inference stage}: The model computes only at the precomputed key timesteps in $\mathcal{T}$.
    Features at the remaining timesteps are efficiently predicted using Eq.~\ref{eq:taylor_pred}.
\end{itemize}
Importantly, DPCache is agnostic to the specific prediction method, as long as the same predictor is consistently used during calibration and inference. 
This design enables seamless integration of future predictors without modifying the path planning framework.

\subsection{Cost Tensor Construction}
A fundamental challenge in global schedule optimization lies in accurately modeling the cost of skipping timesteps.
A straightforward approach would be to define a 2D pairwise cost matrix, where each entry measures the prediction error between two timesteps.
However, in caching-based diffusion sampling with feature prediction, the predicted features depend not only on the current timestep but also on previously cached states.
This path dependency implies that the actual cost of skipping between two timesteps is conditioned on preceding key timesteps, rendering 2D representations insufficient.

To capture this path dependency, we propose the Path-Aware Cost Tensor (PACT), a 3D structure that encodes the skipping cost conditioned on the preceding key timestep.
Formally, PACT is defined as a tensor $\mathcal{C}\in \mathbb{R}^{(T+1)\times (T+1)\times (T+1)}$, where each entry $\mathcal{C}[i, j, k]$ (with $i>j>k$) represents the cumulative cost of skipping from timestep $j$ to $k$, given that $i$ is the preceding key timestep in the sampling path.
To unify boundary handling in subsequent optimal path selection, we introduce a sentinel timestep $t=0$, whose features are predicted from the nearest computed timesteps.
For memory efficiency, we construct PACT using only the final-layer features, \textit{i.e.}, $\mathcal{H}_t=\{\boldsymbol{h}_t^L\}$.
The cost $\mathcal{C}[i,j,k]$ is computed as the cumulative L1 prediction error over all intermediate timesteps skipped when transitioning from $j$ to $k$:
\begin{equation}
    \mathcal{C}[i,j,k] = \sum_{\tau=k}^{j-1}\lVert\boldsymbol{h}_{\tau}^L - \boldsymbol{h}_{pred, \tau}^L(i,j)\rVert_1,
\end{equation}
where $\boldsymbol{h}_{pred, \tau}^L(i,j)$ denotes the predicted feature at timestep $\tau$ using cached features at timesteps $i$ and $j$.
This cumulative formulation ensures that skipping over multiple timesteps incurs the total error across all intermediate predictions, rather than merely the error at the target step.
By aggregating costs over the entire skipped interval, our design discourages large, unstable transitions that may appear locally optimal but degrade global trajectory fidelity.

\begin{algorithm}[t]
\caption{Optimal Schedule Selection via DP}
\label{alg:dp}
\begin{algorithmic}[1]
\Require full steps $T$, target steps $K$, fixed initial steps $M$, path-aware cost tensor $\mathcal{C}[\cdot,\cdot,\cdot]$

\State Initialize $D[m, k] \gets +\infty$, $P[m, k] \gets \text{null}$ for all $m \in [1, K\!+\!1]$, $k \in [0, T]$
\For{$m = 1$ to $M$}\Comment{Fix initial steps}
    \State $k \gets T - m + 1$
    \State $D[m, k] \gets 0$, \quad $P[m, k] \gets k + 1$
\EndFor

\For{$m = M+1$ to $K+1$}\Comment{DP}
    \For{$k = T - m + 1$ downto $0$}
        \State Compute $D[m,k]$ and $P[m,k]$ using Eq.~\ref{eq:dp}
    \EndFor
\EndFor

\State $t_{K+1} \gets 0$
\For{$m = K$ downto $1$}\Comment{Backtracking}
    \State $t_{m} \gets P[m+1, t_{m+1}]$
\EndFor
\State \Return $\mathcal{T} = \{t_1 = T > t_2 > \cdots > t_K > 0\}$
\end{algorithmic}
\end{algorithm}

\subsection{Optimal Schedule Selection}
Given the path-aware cost tensor, our objective is to select an optimal $K$-step sampling schedule $\mathcal{T} = \{ t_1=T > t_2 > \cdots > t_K > 0\}$ that minimizes the total deviation from the full-step denoising trajectory.
To this end, we define the total cost of a schedule $\mathcal{T}$ as the sum of path-aware segment costs, leading to the optimization problem:
\begin{equation}
    \arg\min_{\mathcal{T}} \sum_{m=2}^{K}\mathcal{C}[t_{m-1}, t_m, t_{m+1}],
\end{equation}
where a sentinel endpoint $t_{K+1}=0$ is introduced to provide a fixed terminal reference, ensuring that the cost term involving the last selected step $t_K$ is well-defined.

This path optimization problem can be efficiently solved using dynamic programming (DP).
As summarized in Algorithm~\ref{alg:dp}, we maintain two tables:
\begin{itemize}
    \item DP table $D[m,k]$ stores the minimum cumulative cost to reach timestep $k$ using exactly $m$ key timesteps;
    \item Path table $P[m,k]$ records the predecessor of $k$ in the optimal partial schedule achieving $D[m,k]$, which is subsequently used for backtracking.
\end{itemize}
To preserve critical early-stage denoising dynamics and ensure stable feature prediction, we enforce the first $M$ timesteps to be included in every feasible schedule.
Starting from these fixed initial timesteps, the DP procedure then iteratively extends partial schedules up to the target length $K$.
The recurrence respects the path-dependent structure of PACT by conditioning each transition on its true predecessor, thereby yielding
\begin{equation}
\label{eq:dp}
    D[m,k]=\min_{j>k} D[m-1,j]+\mathcal{C}\bigl[ P\left[m-1,j\right],j,k \bigr],
\end{equation}
where the minimizing $j$ is stored in $P[m,k]$.
The optimization concludes at the terminal state $D[K+1,0]$, and the full schedule $\mathcal{T}$ is recovered by backtracking through $P$.
By casting optimal schedule selection as a dynamic programming problem, we compute the globally optimal schedule exactly and efficiently.
The DP algorithm has time complexity $O(KT^2)$ and space complexity $O(KT)$. 
With $K<T=50$, schedule selection incurs negligible overhead and is performed only once, adding no per-sample latency.



%% file: paper_sec/4_exp.tex
\section{Experiments}
\label{sec: exp}

\subsection{Experimental Settings}

\firstsentence{Model configuration.}
We evaluate the effectiveness of our method on commonly used diffusion models: FLUX.1-dev~\cite{flux2024} for text-to-image generation, HunyuanVideo~\cite{kong2024hunyuanvideo} for text-to-video generation, and DiT-XL/2~\cite{peebles2023scalable} for class-to-image generation.
All experiments are conducted on a single NVIDIA H20 GPU.
For DPCache, we randomly select approximately ten samples as the calibration set for each dataset, set the prediction order to $O=2$ , and fix the first $M=3$ timesteps as mandatory computation steps.
The hyperparameters for other methods in the tables are provided in the supplementary materials.

\firstsentence{Datasets and evaluation metrics.}
For text-to-image generation, we generate 1024$\times$1024 images using 200 DrawBench benchmark~\cite{saharia2022photorealistic} prompts, and evaluate image quality and text-image alignment using ImageReward~\cite{xu2023imagereward} and CLIP Score~\cite{hessel2021clipscore}.
For text-to-video generation, we adopt the VBench evaluation framework~\cite{huang2024vbench} which includes 946 prompts.
We generate five video samples per prompt at a resolution of 640$\times$480$\times$65, resulting in a total of 4,730 videos and assess performance across VBench's 16 core evaluation dimensions. 
For class-to-image generation, we uniformly sample prompts from 1,000 ImageNet~\cite{russakovsky2015imagenet} categories and generate 50,000 images at 256$\times$256 resolution.
We report FID-50k~\cite{heusel2017gans} as the primary metric, complemented by sFID and Inception Score.
Fidelity to the original outputs is evaluated using PSNR, SSIM~\cite{wang2004image}, and LPIPS~\cite{zhang2018unreasonable}, while efficiency is measured by average latency per prompt, FLOPs, and GPU memory usage.

\begin{table*}[htbp]
    \centering
    \caption{Quantitative results of text-to-image generation task for FLUX.1-dev~\cite{flux2024} on DrawBench dataset.}
    \label{tab:flux}
    \adjustbox{max width=\textwidth}{
        \begin{tabular}{l|cc|cc|c|c|c|c|c|c}
           \toprule
           \multirow{2}{*}{\bf Method} & \multicolumn{5}{c|}{\bf Efficiency} & \textbf{Image}$\uparrow$ & \textbf{CLIP}$\uparrow$ & \multirow{2}{*}{\textbf{PSNR}~$\uparrow$} & \multirow{2}{*}{\textbf{SSIM}~$\uparrow$} & \multirow{2}{*}{\textbf{LPIPS}~$\downarrow$} \\
           & \textbf{Latency(s)}~$\downarrow$ & \textbf{Speed}~$\uparrow$ & \textbf{FLOPs(T)}~$\downarrow$ & \textbf{Ratio}~$\uparrow$ & \textbf{Mem(GB)}~$\downarrow$ & \textbf{Reward} & \textbf{Score} & & & \\
           \midrule
           \textbf{50 steps} & 39.54 & 1.00$\times$ & 2990.96 & 1.00$\times$ & 37.41 & 0.979 & 17.40 & - & - & - \\
           \midrule
           \textbf{30}$\mathbf{\%}$ \textbf{steps} & 12.08 & 3.27$\times$ & 907.64 & 3.30$\times$ & 37.41 & 0.855 & 17.12 & 15.32 & 0.6527 & 0.4133 \\
           \textbf{TeaCache}$^1$~\cite{liu2025timestep} & 11.56 & 3.42$\times$ & 862.69 & 3.47$\times$ & 37.45 & 0.934 & 17.17 & 16.31 & 0.6812 & 0.3717 \\
           \textbf{TaylorSeer}$^1$~\cite{liu2025reusing} & 11.28 & 3.51$\times$ & 730.31 & 4.10$\times$ & 43.45 & 0.939 & 17.31 & 16.95 & 0.6922 & 0.3407 \\
           \textbf{SpeCa}$^1$~\cite{liu2025speca} & 10.92 & 3.62$\times$ & 811.82 & 3.68$\times$ & 41.02 & 0.975 & 17.27 & 18.35 & 0.6773 & 0.2920 \\
           \rowcolor{mygray} \textbf{DPCache ($K=13$)} & 11.18 & 3.54$\times$ & 789.50 & 3.79$\times$ & 37.57 & \textbf{1.007} & \textbf{17.34} & \textbf{21.65} & \textbf{0.8106} & \textbf{0.1804} \\
           \midrule
           \textbf{22}$\mathbf{\%}$ \textbf{steps} & 8.94 & 4.42$\times$ & 669.54 & 4.47$\times$ & 37.41 & 0.786 & 17.07 & 14.49 & 0.6138 & 0.4717 \\
           \textbf{TeaCache}$^2$~\cite{liu2025timestep} & 8.59 & 4.60$\times$ & 638.35 & 4.69$\times$ & 37.45 & 0.864 & 17.14 & 15.29 & 0.6361 & 0.4451 \\
           \textbf{TaylorSeer}$^2$~\cite{liu2025reusing} & 8.27 & 4.78$\times$ & 492.34 & 6.07$\times$ & 43.45 & 0.793 & 17.22 & 14.76 & 0.5782 & 0.4929 \\
           \textbf{SpeCa}$^2$~\cite{liu2025speca} & 8.54 & 4.63$\times$ & 625.57 & 4.78$\times$ & 41.02 & 0.927 & 17.23 & 16.34 & 0.6017 & 0.4063 \\
           \rowcolor{mygray} \textbf{DPCache ($K=9$)} & 8.12 & 4.87$\times$ & 551.50 & 5.42$\times$ & 37.57 & \textbf{0.958} & \textbf{17.33} & \textbf{18.77} & \textbf{0.7117} & \textbf{0.2946} \\
           \bottomrule
        \end{tabular}
    }
\end{table*}

\subsection{Comparison with other methods}
\label{sec:compare_with_others}
\subsubsection{Text-to-Image Generation}
\firstsentence{Quantitative results.}
Table~\ref{tab:flux} presents the quantitative results of various acceleration methods applied to FLUX.1-dev~\cite{flux2024} on the DrawBench~\cite{saharia2022photorealistic} dataset.
Compared to existing approaches, our proposed DPCache achieves the best performance across all evaluation metrics under comparable speedup ratios, and even attains performance competitive with the unaccelerated baseline ($+$0.028 in ImageReward) under 3.54$\times$ acceleration.
Under a more aggressive 4.87$\times$ speedup, DPCache surpasses the second-best method by 0.031 in ImageReward and by 0.10 in CLIP Score, demonstrating superior generation quality and stronger text–image alignment. 
Moreover, on the metrics that measure fidelity to the unaccelerated baseline, DPCache outperforms other methods by a large margin, with improvements of $+$2.43 in PSNR, $+$0.11 in SSIM, and $-$0.11 in LPIPS, strongly validating that the globally optimal schedule identified by DPCache closely approximates the original sampling trajectory, effectively minimizing deviation from the unaccelerated generation process.

\begin{figure}[t]
    \centering
    \includegraphics[width=\linewidth]{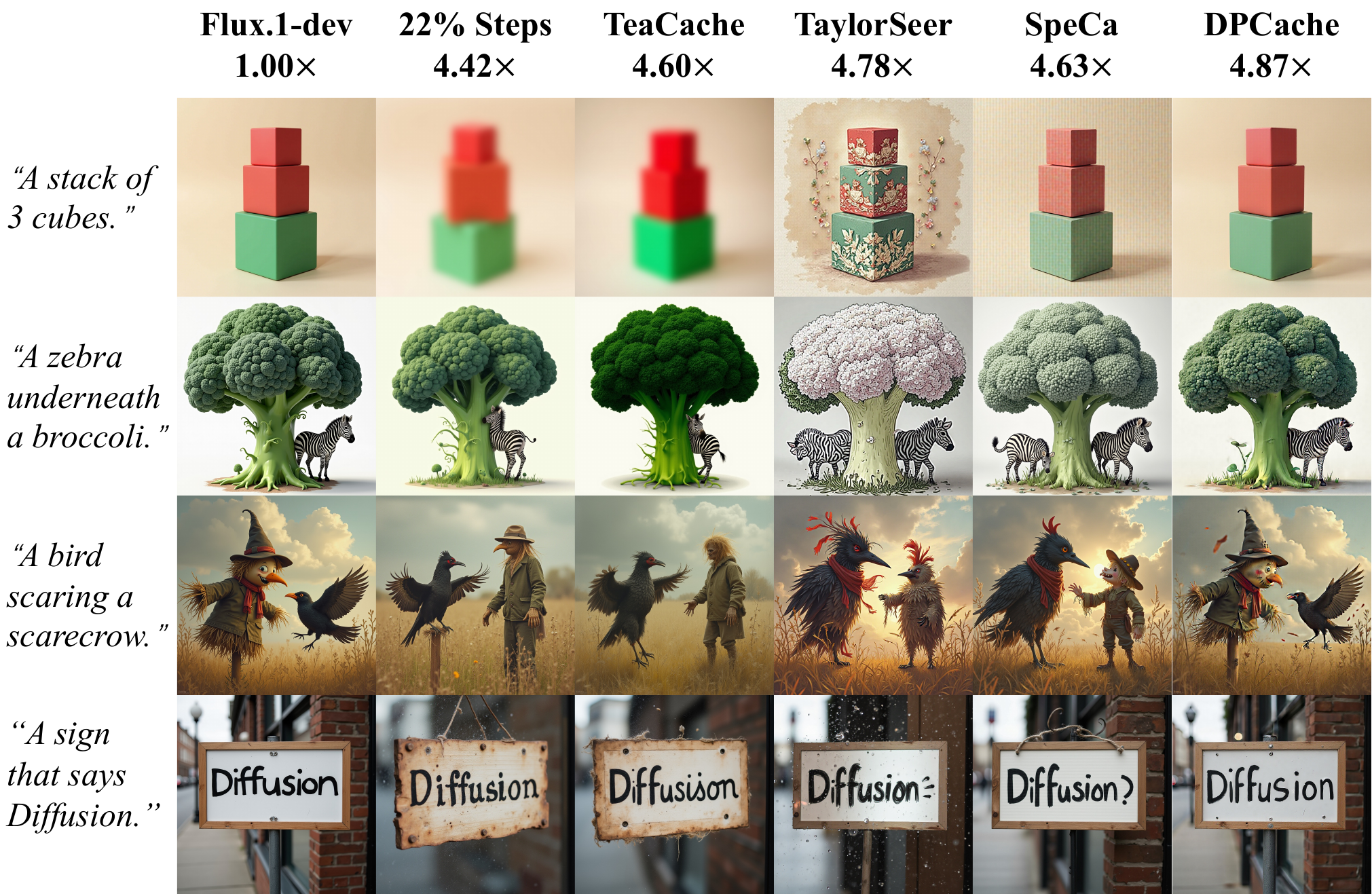}
    \caption{Qualitative comparison of different acceleration methods applied to FLUX.1-dev on DrawBench dataset.}
    \label{fig:vis_flux}
    \vspace{-0.1cm}
\end{figure}

\firstsentence{Qualitative results.}
Figure~\ref{fig:vis_flux} presents the qualitative comparison of different acceleration methods on FLUX.1-dev.
In the first row, directly reducing sampling steps or using TeaCache leads to overall blurriness. 
Prediction-based methods like TaylorSeer and SpeCa avoid blurriness, but TaylorSeer's fixed schedule causes its output to deviate significantly from the baseline, while SpeCa’s locally adaptive strategy introduces noise in the background. 
In contrast, DPCache produces sharp edges and a clean background, closely matching the full-step baseline, which demonstrates the effectiveness of our global scheduling approach.
Similar trends are observed in the second row, where TeaCache exhibits structural distortions in the zebra, while TaylorSeer and SpeCa generate two zebras, which is inconsistent with the prompt.
Moreover, rows 3 and 4 show that DPCache excels in complex scene generation and text rendering, preserving layout and context while maintaining high fidelity.

\begin{table*}[htbp]
    \centering
    \caption{Quantitative results of text-to-video generation task for HunyuanVideo~\cite{kong2024hunyuanvideo} on VBench dataset.}
    \label{tab:hunyuan}
    \adjustbox{max width=\textwidth}{
        \begin{tabular}{l|cc|cc|c|c|c|c|c}
           \toprule
           \multirow{2}{*}{\bf Method} & \multicolumn{5}{c|}{\bf Efficiency} & \textbf{VBench}$\uparrow$ &  \multirow{2}{*}{\textbf{PSNR}~$\uparrow$} & \multirow{2}{*}{\textbf{SSIM}~$\uparrow$} & \multirow{2}{*}{\textbf{LPIPS}~$\downarrow$} \\
           & \textbf{Latency(s)}~$\downarrow$ & \textbf{Speed}~$\uparrow$ & \textbf{FLOPs(T)}~$\downarrow$ & \textbf{Ratio}~$\uparrow$ & \textbf{Mem(GB)}~$\downarrow$ & \textbf{Score(\%)} & & & \\
           \midrule
           \textbf{50 steps} & 332.92 & 1.00$\times$ & 14050.37 & 1.00$\times$ & 60.22 & 80.93 & - & - & - \\
           \midrule
           \textbf{22}$\mathbf{\%}$ \textbf{steps} & 82.47 & 4.04$\times$ & 3095.22 & 4.54$\times$ & 60.22 & 78.89 & 16.97 & 0.5753 & 0.4654 \\
           \textbf{TeaCache}$^1$~\cite{liu2025timestep} & 79.56 & 4.18$\times$ & 2972.71 & 4.73$\times$ & 60.58 & 79.25 & 17.55 & 0.5801 & 0.4081 \\
           \textbf{TaylorSeer}$^1$~\cite{liu2025reusing} & 86.11 & 3.87$\times$ & 2821.20 & 4.98$\times$ & 84.47 & 80.33 & 18.53 & 0.6025 & 0.3722 \\
           \textbf{SpeCa}$^1$~\cite{liu2025speca} & 82.15 & 4.05$\times$ & 2731.61 & 5.14$\times$ & 84.47 & 80.26 & 20.09 & 0.6370 & 0.3554 \\
           \rowcolor{mygray} \textbf{DPCache ($K=11$)} & 83.22 & 4.05$\times$ & 3100.33 & 4.53$\times$ & 60.58 & \textbf{80.35} & \textbf{23.11} & \textbf{0.7462} & \textbf{0.1890} \\
           \midrule
           \textbf{TeaCache}$^2$~\cite{liu2025timestep} & 70.05 & 4.75$\times$ & 2538.79 & 5.53$\times$ & 60.58 & 78.17 & 17.75 & 0.5861 & 0.4137 \\
           \textbf{TaylorSeer}$^2$~\cite{liu2025reusing} & 79.84 & 4.17$\times$ & 2540.46 & 5.53$\times$ & 84.47 & 79.99 & 18.23 & 0.5899 & 0.3967 \\
           \textbf{SpeCa}$^2$~\cite{liu2025speca} & 72.09 & 4.62$\times$ & 2254.75 & 6.23$\times$ & 84.47 & 78.50 & 17.83 & 0.5867 & 0.4085 \\
           \rowcolor{mygray} \textbf{DPCache ($K=9$)} & 70.09 & 4.75$\times$ & 2538.79 & 5.53$\times$ & 60.58 & \textbf{80.23} & \textbf{21.04} & \textbf{0.6852} & \textbf{0.2504} \\
           \bottomrule
        \end{tabular}
    }
\end{table*}

\begin{figure*}
    \centering
    \includegraphics[width=\linewidth]{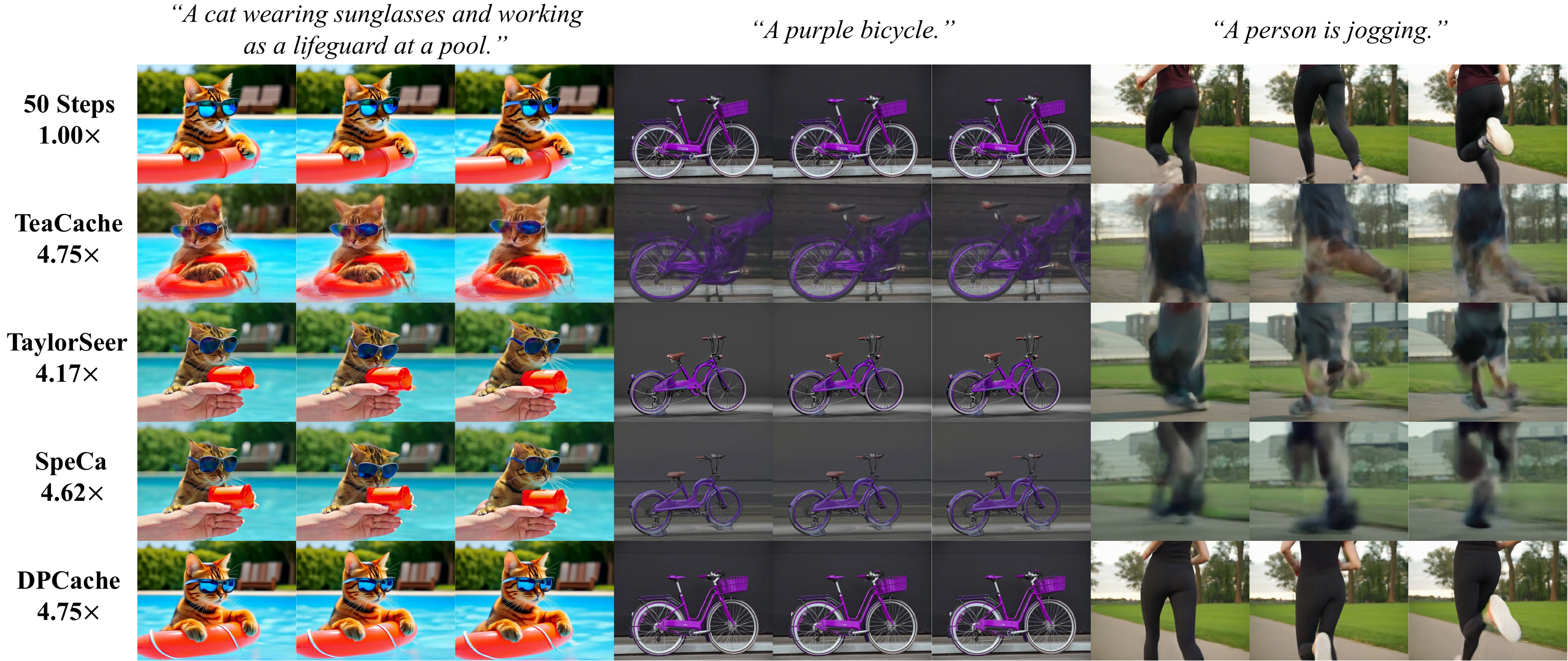}
    \caption{Qualitative comparison of different acceleration methods applied to HunyuanVideo on VBench dataset.}
    \label{fig:vis_hunyuan}
\end{figure*}
    
\subsubsection{Text-to-Video Generation}
\firstsentence{Quantitative results.}
We further validate the effectiveness of our method on text-to-video generation, a more challenging setting involving larger models, as shown in Table~\ref{tab:hunyuan}.
Under a 4.75$\times$ speedup, DPCache achieves a VBench score of 80.23$\%$, substantially outperforming existing methods by $+$0.24$\%$.
In particular, DPCache demonstrates significantly higher fidelity to the original outputs across all fidelity-related metrics.
Notably, existing prediction-based approaches such as TaylorSeer and SpeCa require caching intermediate features from all transformer layers, resulting in considerable GPU memory overhead ($+$24.25 GB). In contrast, DPCache caches features only from the final layer, incurring only a negligible increase ($+$0.36 GB) in memory usage compared to the baseline.
This minimal memory overhead confirms that DPCache remains practical and scalable in the era of large-scale generative models.

\firstsentence{Qualitative results.}
Figure~\ref{fig:vis_hunyuan} showcases the qualitative comparison of different acceleration methods on HunyuanVideo.
In the first column, TeaCache distorts the lifebuoy's shape, while TaylorSeer and SpeCa misinterpret it entirely as a human hand, indicating a severe deviation from the baseline's denoising trajectory.
In contrast, DPCache accurately retains the lifebuoy with only minor visual differences from the reference.
In the second column, all other acceleration methods introduce blur or geometric warping, whereas DPCache delivers sharp edges and accurate structure that closely match the full-step output.
In the third column, existing methods suffer from heavy motion blur, and even the baseline exhibits slight artifacts in a few frames.
However, DPCache maintains both visual sharpness and smooth motion across the entire sequence, demonstrating superior stability in high-dynamic scenarios.

\begin{table*}[htbp]
    \centering
    \caption{Quantitative results of class-to-image generation task for DiT-XL/2~\cite{peebles2023scalable} on ImageNet dataset.}
    \label{tab:dit}
    \adjustbox{max width=\linewidth}{
        \begin{tabular}{l|cc|cc|c|cc|c}
           \toprule
           \textbf{Method} & \textbf{Latency(s)}~$\downarrow$ &\textbf{Speed}~$\uparrow$ & \textbf{FLOPs(T)}~$\downarrow$ & \textbf{Ratio}~$\uparrow$ & \textbf{Mem(GB)}~$\downarrow$ & \textbf{FID}~$\downarrow$ & \textbf{sFID}~$\downarrow$ & \textbf{Inception Score}~$\uparrow$  \\
           \midrule
           \textbf{DDIM-50 steps} & 0.851 & 1.00$\times$ & 22.88 & 1.00$\times$ & 3.183 & 2.226 & 4.275 & 241.10 \\
           \midrule
           \textbf{40}$\mathbf{\%}$ \textbf{steps} & 0.341 & 2.50$\times$ & 9.154 & 2.50$\times$ & 3.183 & 3.504 & 4.918 & 221.45 \\
           \textbf{TaylorSeer}$^1$~\cite{liu2025reusing} & 0.444 & 1.92$\times$ & 3.256 & 7.03$\times$ & 3.278 & 4.776 & 7.793 & 201.94 \\
           \textbf{SpeCa}$^1$~\cite{liu2025speca} & 0.384 & 2.22$\times$ & 4.285 & 5.34$\times$ & 3.282 & 3.185 & 5.690 & 224.24 \\
           \rowcolor{mygray} \textbf{DPCache ($K=16$)} & 0.344 & 2.47$\times$ & 7.326 & 3.12$\times$ & 3.190 & \textbf{2.562} & \textbf{4.503} & \textbf{230.60} \\
           \midrule
           \textbf{30}$\mathbf{\%}$ \textbf{steps} & 0.257 & 3.31$\times$ & 6.865 & 3.33$\times$ & 3.183 & 5.138 & 6.120 & 205.65 \\
           \textbf{TaylorSeer}$^2$~\cite{liu2025reusing} & 0.431 & 1.97$\times$ & 2.800 & 8.17$\times$ & 3.278 & 6.163 & 9.503 & 186.55 \\
           \textbf{SpeCa$^2$}~\cite{liu2025speca} & 0.367 & 2.32$\times$ & 3.656 & 6.26$\times$ & 3.282 & 4.705 & 7.835 & 203.17 \\
           \rowcolor{mygray} \textbf{DPCache ($K=12$)} & 0.282 & 3.02$\times$ & 5.495 & 4.16$\times$ & 3.190 & \textbf{3.285} & \textbf{5.063} & \textbf{215.99} \\
           \bottomrule
        \end{tabular}
    }
\end{table*}

\subsubsection{Class-to-Image Generation}
As shown in Table~\ref{tab:dit}, we also evaluate DPCache on class-conditional image generation using the DiT-XL/2~\cite{peebles2023scalable}.
Due to its small parameter count and low output resolution, acceleration gains are inherently limited for all methods. 
Although TaylorSeer and SpeCa achieve lower FLOPs by reusing cached features, they store intermediate activations from all transformer layers, incurring substantial memory bandwidth overhead that dominates actual latency.
In contrast, DPCache caches only the final-layer features, thereby reducing both computation and memory pressure.
DPCache achieves a 3.02$\times$ speedup while delivering the best generation quality among all approaches, with an FID of 3.285, sFID of 5.063, and Inception Score of 215.99. 
DPCache outperforms the second-best method by $-$1.420 in FID, $-$2.772 in sFID, and $+$10.34 in Inception Score, demonstrating its effectiveness across diverse model scales.

\subsection{Ablation Studies}
\firstsentence{Path-aware cost tensor.}
As shown in Table~\ref{tab:abl_pact}, we conduct ablation studies to evaluate the impact of PACT’s dimensionality and cost aggregation strategy.
The first row adopts a 2D cost that evaluates only the prediction error at the current timestep. Although it ignores the influence of previously cached timesteps on the cost, this naive strategy still achieves competitive performance against existing methods, underscoring the core advantage of formulating acceleration as a global path planning problem.
In the second row, applying cumulative error to the 2D cost degrades performance, as the 2D representation inherently misestimates per-step errors, and aggregating these inaccuracies distorts the cost landscape, misleading schedule selection.
The third row leverages the more accurate 3D path-aware cost, improving fidelity to the full-step baseline. 
However, by relying solely on the current-step error, the scheduler selects aggressive steps that appear locally optimal but skip structurally critical timesteps, compromising trajectory stability.
The fourth row integrates 3D cost with cumulative error, capturing both dependence on prior cached states and total fidelity loss over skipped intervals, and yields a $+$0.006 gain in ImageReward and a $+$0.78 improvement in PSNR over the first row, confirming the effectiveness of PACT.


\begin{table}[t]
    \centering
    \caption{Ablation studies of Path-Aware Cost Tensor on FLUX.1-dev with $K=13$.}
    \label{tab:abl_pact}
    \adjustbox{max width=\linewidth}{
        \begin{tabular}{cc|c|c|c|c|c}
           \toprule
           \multicolumn{2}{c|}{\bf Method} & \textbf{Image}$\uparrow$ & \textbf{CLIP}$\uparrow$ & \multirow{2}{*}{\textbf{PSNR}~$\uparrow$} & \multirow{2}{*}{\textbf{SSIM}~$\uparrow$} & \multirow{2}{*}{\textbf{LPIPS}~$\downarrow$} \\
           \textbf{Dim} & \textbf{Sum} & \textbf{Reward} & \textbf{Score} & & & \\
           \midrule
           2D & \ding{56} & 1.001 & 17.33 & 20.87 & 0.7881 & 0.2066 \\
           2D & \ding{52} & 0.977 & 17.34 & 19.46 & 0.7605 & 0.2423 \\
           3D & \ding{56} & 0.998 & 17.33 & 21.05 & 0.7952 & 0.2037 \\
           \rowcolor{mygray} 3D & \ding{52} & \textbf{1.007} & \textbf{17.34} & \textbf{21.65} & \textbf{0.8106} & \textbf{0.1804} \\
           \bottomrule
        \end{tabular}
    }
\end{table}

\firstsentence{Calibration set.}
To assess the impact of calibration set selection on acceleration performance, we conduct an ablation study on the size, sampling strategy, and source of the calibration set, as shown in Table~\ref{tab:abl_cali}.
The first three rows vary the number of calibration samples. 
Remarkably, using 11 or 5 samples yields identical sampling schedules and thus identical performance, while even a single sample achieves competitive results with an ImageReward of 1.001, demonstrating that DPCache requires minimal calibration overhead.
In the fourth row, we replace random sampling with stratified sampling by selecting one prompt from each of the 11 DrawBench categories. 
The resulting performance remains unchanged, indicating that careful selection of calibration prompts is unnecessary.
In the fifth row, we further switch the calibration source to PartiPrompts~\cite{yu2022scaling}, a dataset that differs substantially from DrawBench in both semantics and linguistic style. 
Despite this distribution shift, the selected schedule and generation quality are preserved.
These findings strongly validate the core assumption in Section~\ref{sec:overview} that the denoising trajectory shape is largely content-agnostic, and they confirm the robustness of DPCache to the choice of the calibration set.

%% file: paper_sec/5_conclusion.tex
\section{Conclusion}
\label{sec: conclusion}

In this paper, We propose DPCache, a training-free diffusion acceleration framework that formulates sampling as a global path planning problem. 
By constructing a Path-Aware Cost Tensor (PACT) from a few calibration samples, DPCache captures the trajectory-dependent error of skipping timesteps conditioned on prior key steps. 
Using dynamic programming, it selects an optimal sequence of key timesteps that minimizes total deviation from the full denoising trajectory. 
During inference, only these key steps require full computation, while intermediate steps are efficiently predicted using cached features. 
Extensive experiments on DiT, FLUX, and HunyuanVideo show DPCache achieves up to 4.87$\times$ speedup with consistently better generation quality than prior caching-based methods, demonstrating the effectiveness of the path-aware global scheduling framework.
Future work includes extending DPCache to input-adaptive scheduling and integrating learnable predictors for more accurate feature reconstruction.

\begin{table}[t]
    \centering
    \caption{Ablation study on calibration set size, sampling strategy, and prompt source on FLUX.1-dev with $K=13$.}
    \label{tab:abl_cali}
    \adjustbox{max width=\linewidth}{
        \begin{tabular}{ccc|c|c|c}
           \toprule
           \multirow{2}{*}{\bf Num} & \multirow{2}{*}{\bf Strategy} & \multirow{2}{*}{\bf Source} & \textbf{Image}$\uparrow$ & \textbf{CLIP}$\uparrow$ & \multirow{2}{*}{\textbf{PSNR}~$\uparrow$} \\
           & & & \textbf{Reward} & \textbf{Score} & \\
           \midrule
           11 & Random & DrawBench & 1.007 & 17.34 & 21.65 \\
           5 & Random & DrawBench & 1.007 & 17.34 & 21.65 \\
           1 & Random & DrawBench & 1.001 & 17.33 & 21.64 \\
           11 & Stratified & DrawBench & 1.007 & 17.34 & 21.65 \\
           11 & Random & PartiPrompts & 1.007 & 17.34 & 21.65 \\
           \bottomrule
        \end{tabular}
    }
\end{table}


%% file: paper_sec/appendix.tex
\appendix
\section{Experimental Settings}
To ensure reproducibility and fair comparison across methods, Table~\ref{tab:hyper} provides a detailed overview of the hyperparameter settings used for each method in Section~4.2 of the main paper.

\begin{table}[h]
    \centering
    \caption{Hyperparameters used for each method in our experiments.}
    \begin{tabular}{c|c|l}
        \toprule
        \textbf{Model} & \textbf{Method} & \textbf{Hyperparameters} \\
        \midrule
        \multirow{6}{*}{FLUX.1-dev} & TeaCache$^1$ & $l=0.7$ \\
        & TeaCache$^2$ & $l=1.0$ \\
        & TaylorSeer$^1$ & $\mathcal{N}=5, O=2$ \\
        & TaylorSeer$^2$ & $\mathcal{N}=8, O=2$ \\
        & SpeCa$^1$ & $\tau=1.0, \beta=0.2$ \\
        & SpeCa$^2$ & $\tau=8.5, \beta=0.4$ \\
        \midrule
        \multirow{6}{*}{HunyuanVideo} & TeaCache$^1$ & $l=0.4$ \\
        & TeaCache$^2$ & $l=0.5$ \\
        & TaylorSeer$^1$ & $\mathcal{N}=5, O=1$ \\
        & TaylorSeer$^2$ & $\mathcal{N}=8, O=1$ \\
        & SpeCa$^1$ & $\tau=1.0, \beta=0.1$ \\
        & SpeCa$^2$ & $\tau=2.0, \beta=0.1$ \\
        \midrule
        \multirow{4}{*}{DiT-XL/2} & TaylorSeer$^1$ & $\mathcal{N}=9, O=2$ \\
        & TaylorSeer$^2$ & $\mathcal{N}=10, O=2$ \\
        & SpeCa$^1$ & $\tau=1.0, \beta=0.1$ \\
        & SpeCa$^2$ & $\tau=2.0, \beta=0.3$ \\
        \bottomrule
    \end{tabular}
    
    \label{tab:hyper}
\end{table}

\section{Comparison with State-of-the-Art Methods}
\subsection{Analysis of Sampling Trajectories}
\begin{figure}[t]
    \centering
    \includegraphics[width=\linewidth]{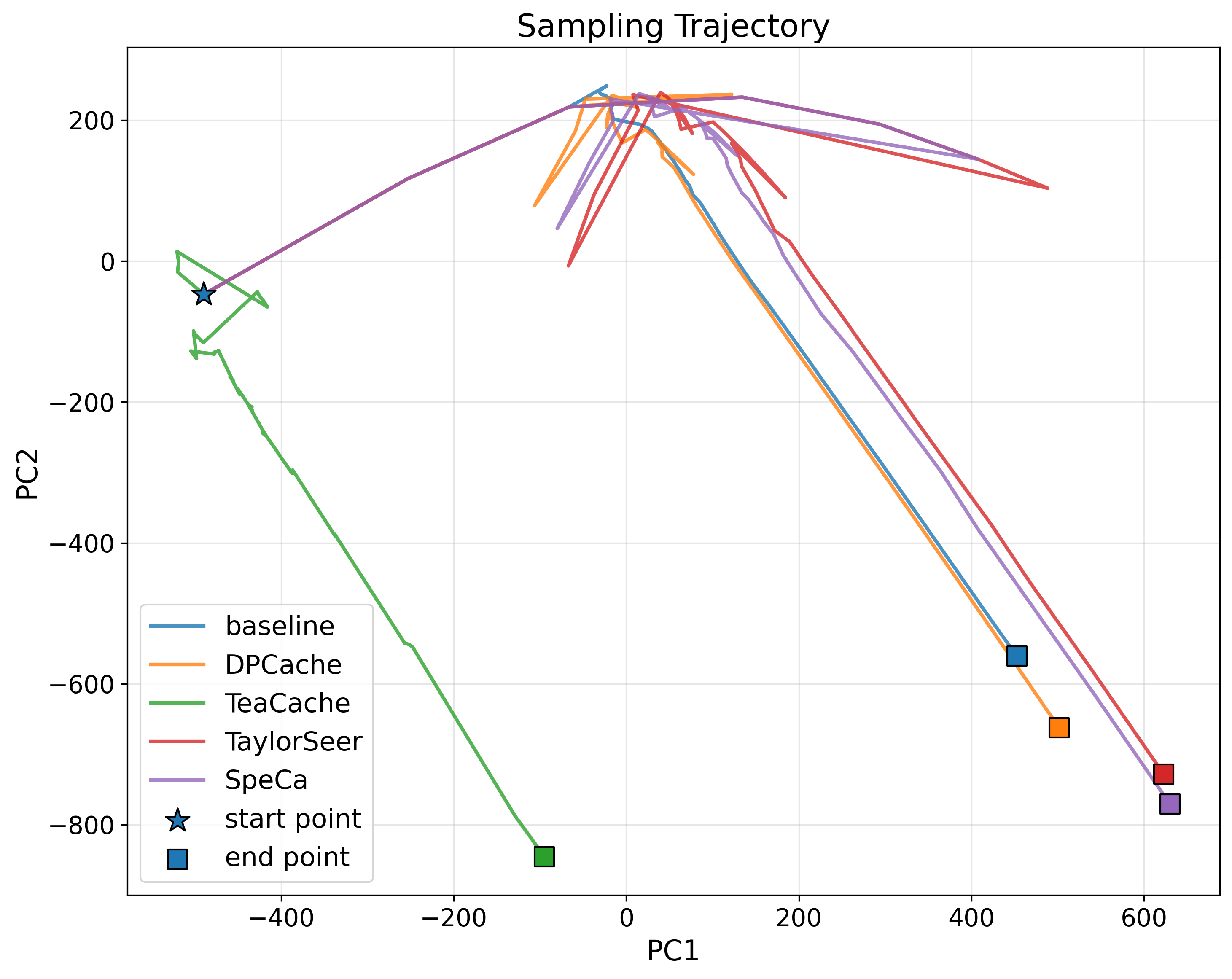}
    \caption{Visualization of sampling trajectories when applying different acceleration methods to FLUX.1-dev.}
    \label{fig:vis_traj}
\end{figure}

To visually demonstrate the superiority of our method, we visualize the sampling trajectories of FLUX.1-dev under different acceleration strategies, as shown in Figure~\ref{fig:vis_traj}.
Specifically, we run 50 samples, compute the average output features at each timestep, apply PCA to project them into a two-dimensional space, and visualize the resulting trajectories in a 2D coordinate system.
DPCache’s trajectory closely follows that of the baseline, confirming that our method minimizes global trajectory deviation, which is consistent with Figure 1 of the main paper. 
Moreover, we notice that prediction timesteps often deviate significantly from the baseline trajectory, but subsequent computation steps effectively steer the trajectory back toward it. 
By selecting computation timesteps from a global perspective, DPCache ensures that the accelerated model remains consistently close to the original sampling trajectory throughout inference. 

\subsection{Quantitative Results}
\begin{figure*}[t]
    \centering
    \begin{minipage}{0.49\linewidth}
        \centering
        \includegraphics[width=\linewidth]{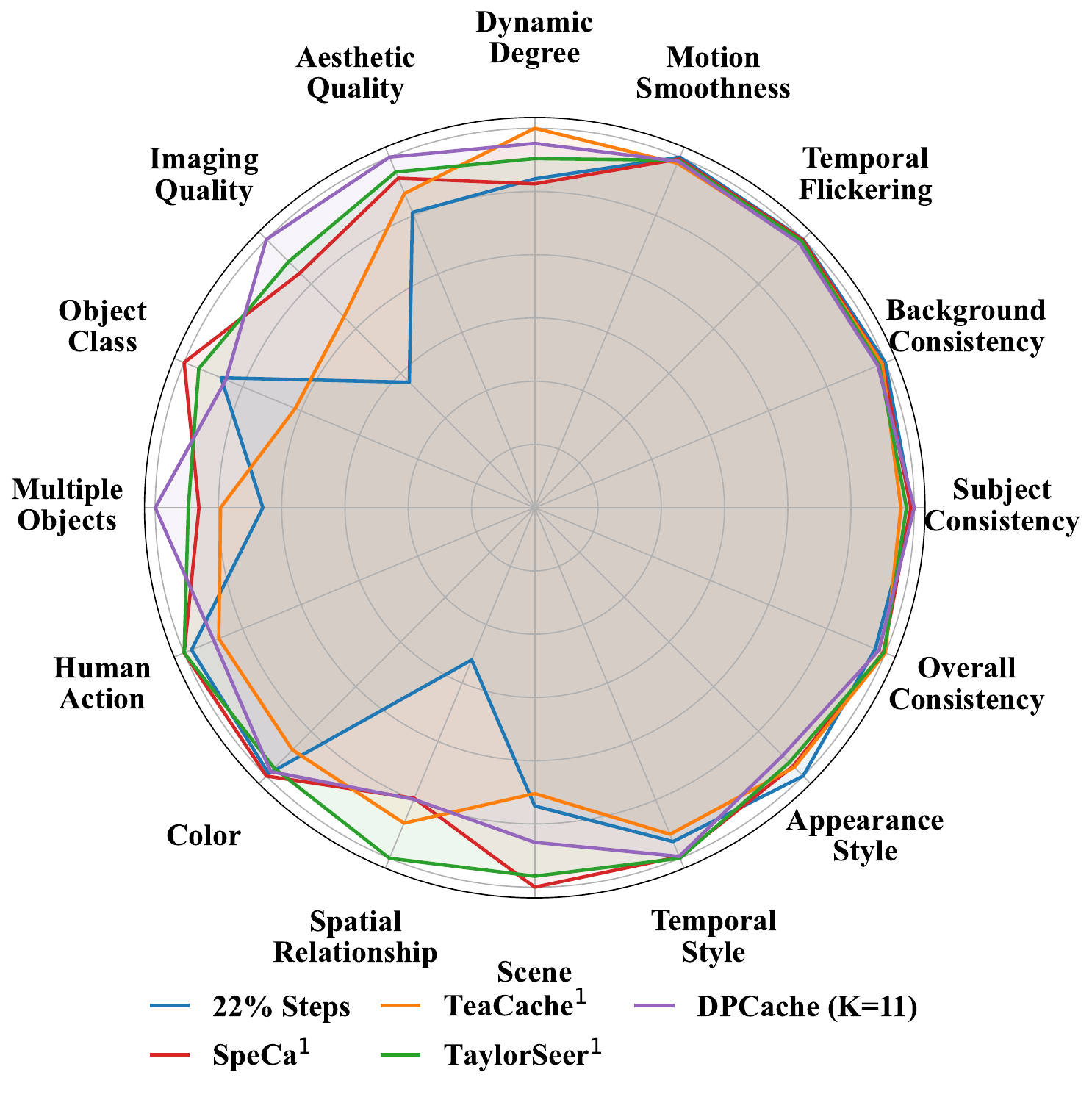}
    \end{minipage}
    \hfill
    \begin{minipage}{0.49\linewidth}
        \centering
        \includegraphics[width=\linewidth]{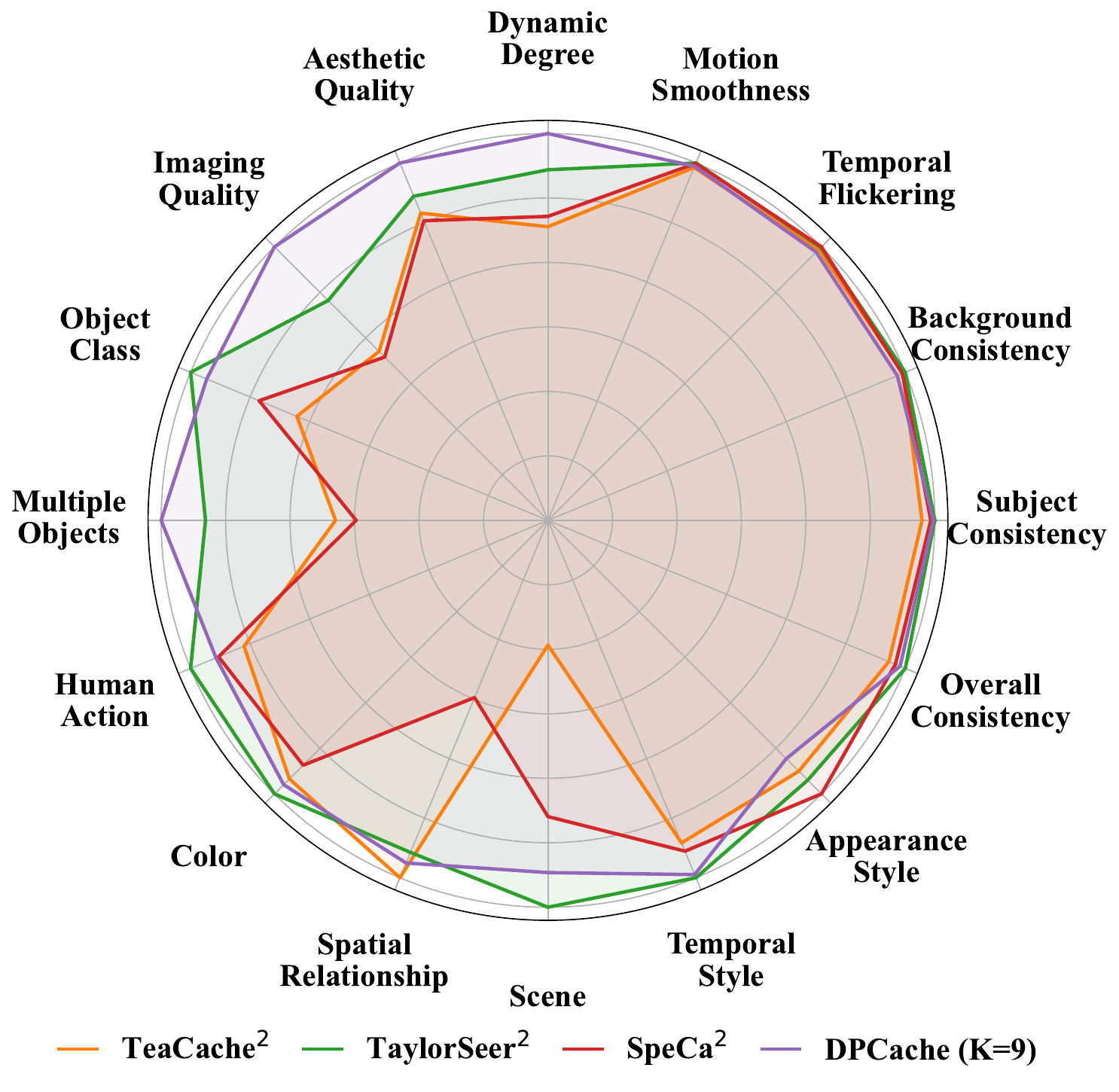}
    \end{minipage}
    \caption{
        VBench performance of HunyuanVideo across all dimensions with various acceleration methods under different acceleration settings: 
        (left) low speedup, (right) high speedup.
        Scores are normalized per dimension for improved visual comparison.
    }
    \label{fig:radar}
    \vspace{-0.1cm}
\end{figure*}

\firstsentence{VBench Score for all dimensions.}
Figure~\ref{fig:radar} presents a comprehensive comparison of HunyuanVideo’s performance across all VBench dimensions with various acceleration methods. Under different acceleration settings, our proposed DPCache consistently achieves competitive performance across all dimensions, and notably outperforms other methods in aesthetic quality, imaging quality, and multiple objects.
Our method demonstrates more pronounced advantages under high-speedup settings, highlighting the potential of our globally optimized sampling schedule to maintain quality while enabling aggressive acceleration.

\subsection{Qualitative Results}
\begin{figure}[t]
    \centering
    \includegraphics[width=\linewidth]{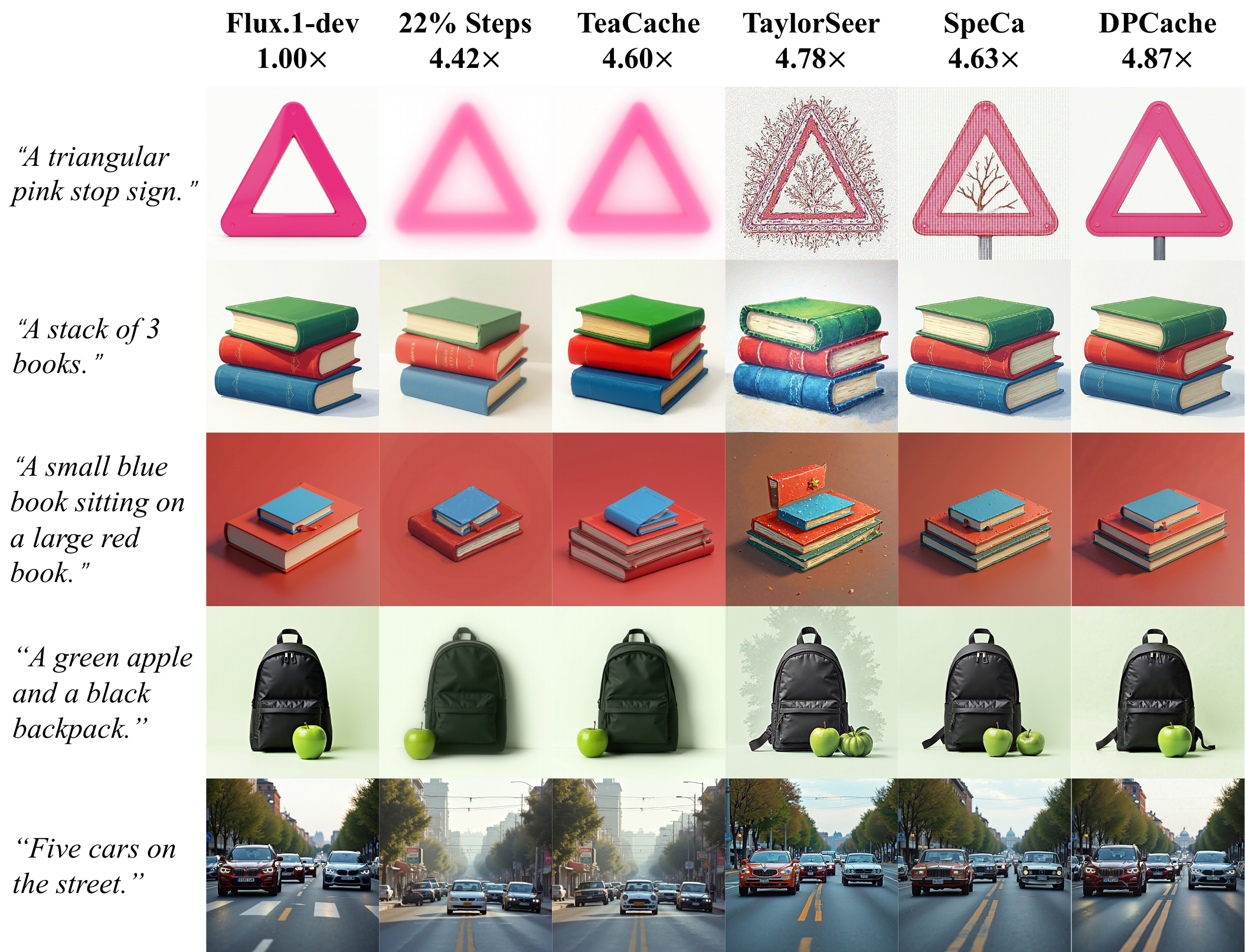}
    \caption{Additional qualitative comparisons of different acceleration methods applied to FLUX.1-dev on DrawBench dataset.}
    \label{fig:flux_more}
    \vspace{-0.3cm}
\end{figure}

\firstsentence{Text-to-image generation.}
We provide additional qualitative comparisons of various acceleration methods applied to FLUX.1-dev on the DrawBench dataset in Figure~\ref{fig:flux_more}.
Existing approaches often exhibit noticeable artifacts, such as blurriness, geometric distortions, or misalignment with the input text prompt, whereas our proposed DPCache maintains high visual fidelity and closely resembles the output of the unaccelerated baseline.

\begin{figure*}
    \centering
    \includegraphics[width=\linewidth]{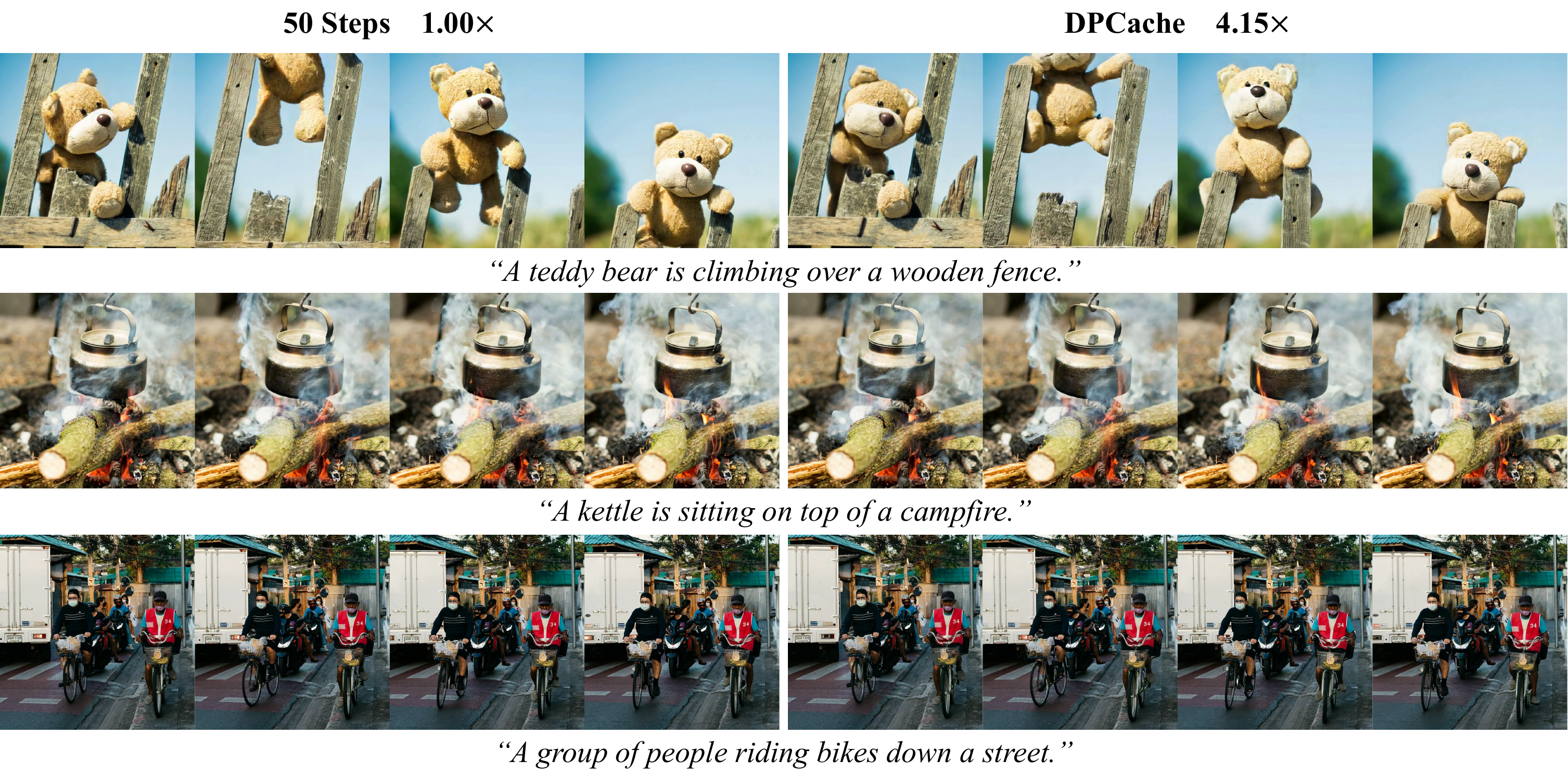}
    \caption{Qualitative comparison between the original (unaccelerated) and DPCache-accelerated Wan-2.1-I2V-14B on the VBench-I2V benchmark.}
    \label{fig:vis_wan}
    \vspace{-0.1cm}
\end{figure*}

\firstsentence{Text-to-video generation.}
To further validate the effectiveness of DPCache, we apply it to Wan-2.1-I2V-14B~\cite{wan2025}, which is a larger, higher-resolution model with state-of-the-art performance in video generation. 
As shown in Figure~\ref{fig:vis_wan}, we sample some inputs from VBench-I2V~\cite{huang2024vbench++} benchmark and compare the generation results of the baseline and DPCache at a resolution of 832$\times$832$\times$81.
DPCache achieves 4.15$\times$ speedup while preserving visual fidelity: the accelerated output maintains sharpness and detail, even in complex scenes, without introducing noticeable artifacts.

\begin{figure}[t]
    \centering
    \includegraphics[width=\linewidth]{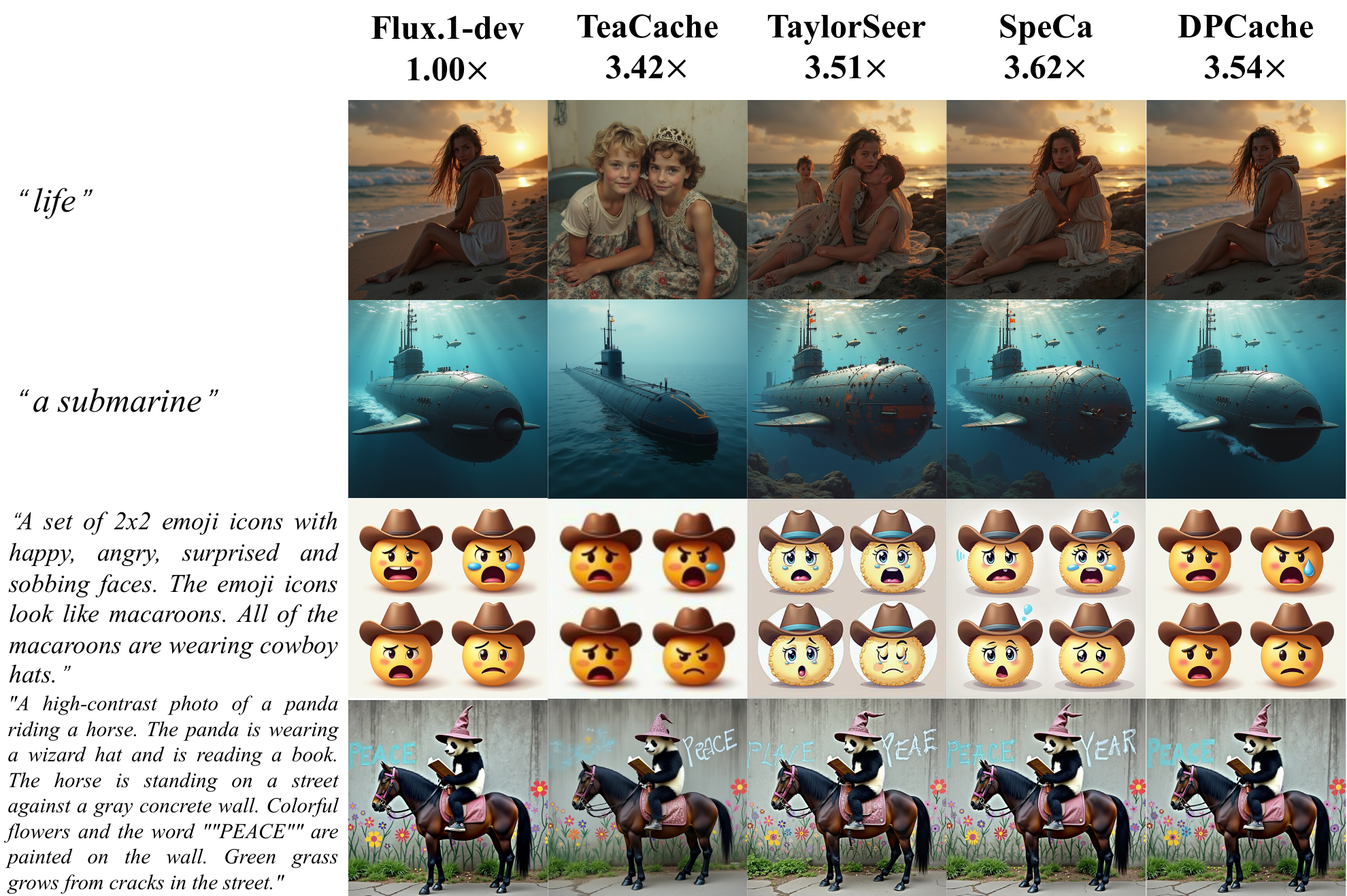}
    \caption{Qualitative comparison of different acceleration methods applied to FLUX.1-dev on out-of-distribution prompts.}
    \label{fig:flux_ood}
    \vspace{-0.3cm}
\end{figure}

\begin{figure}[t]
    \centering
    \includegraphics[width=\linewidth]{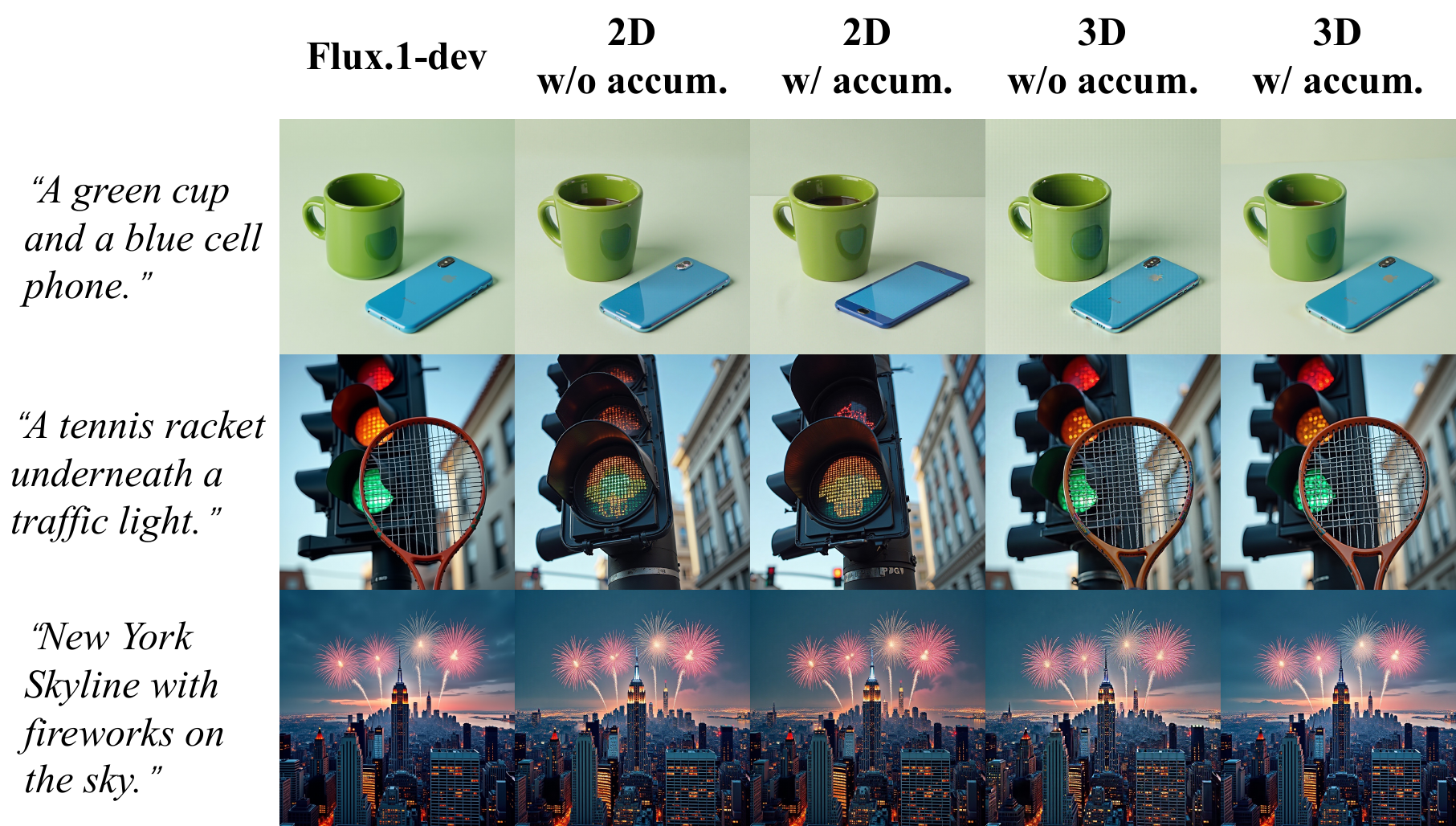}
    \caption{Visualization of ablation studies on the dimensionality and cost aggregation strategy of PACT.}
    \label{fig:vis_abl}
\end{figure}

\firstsentence{Robustness on out-of-distribution prompts.}
In Section 4.3 of the main paper, we quantitatively demonstrate that DPCache is robust to the choice and size of the calibration set. 
To further validate this robustness, we apply a FLUX.1-dev model calibrated on 10 samples from DrawBench, and evaluate it on out-of-distribution prompts from PartiPrompts~\cite{yu2022scaling} that exhibit a significantly different distribution from DrawBench.
As shown in Figure~\ref{fig:flux_ood}, we compare different acceleration methods on out-of-distribution prompts.

In the first row, the prompt is short and abstract. 
The result generated by TeaCache deviates significantly from the baseline, indicating a complete drift from the original sampling trajectory. 
TaylorSeer and SpeCa produce scenes that are semantically similar to the baseline but introduce spurious human figures and distorted structures. 
In contrast, DPCache reproduces a result nearly identical to the baseline.
In the fourth row, the prompt is highly detailed and specifies numerous visual elements. 
TeaCache and TaylorSeer fail to preserve the text on the wall, while SpeCa generates noticeable artifacts on the wizard hat. 
DPCache, however, maintains the finest details with the highest fidelity.

These results on out-of-distribution inputs strongly demonstrate the robustness of DPCache to calibration set selection, confirming that our method can generalize well even when calibrated on a small and distributionally mismatched set.

\section{Ablation Studies}
\firstsentence{Path-aware cost tensor.}
In Section 4.3 of the main paper, we quantitatively validate the effectiveness of the PACT design. Figure~\ref{fig:vis_abl} further visualizes results under different cost tensor dimensionalities and cost aggregation strategies.

In the first row, the result using a 2D cost tensor without temporal accumulation exhibits noticeable distortions: the cup shape and the logo on the phone differ from the baseline result. 
When temporal accumulation is applied to the 2D cost, these errors are further amplified, leading to an even greater deviation in the phone’s appearance. 
In contrast, both variants of the 3D cost tensor achieve high fidelity to the baseline, demonstrating that modeling the influence of the previous key timestep during cost computation better captures the true error of timestep skipping.
Moreover, the non-cumulative variants produce outputs with coarser textures and reduced detail. This aligns with our analysis that non-aggregated costs tend to trigger aggressive skipping in the final stages of sampling, impairing fine-grained synthesis. 
Similar trends are observed in the second and third rows: 2D cost tensors yield lower visual similarity to the baseline, and non-aggregated strategies suffer from significant detail loss.

These results confirm that leveraging a 3D cost tensor with temporal aggregation enables DPCache to estimate skipping-induced error more accurately, thereby preserving structural consistency and fine details during accelerated sampling.

\begin{figure}[t]
    \centering
    \includegraphics[width=\linewidth]{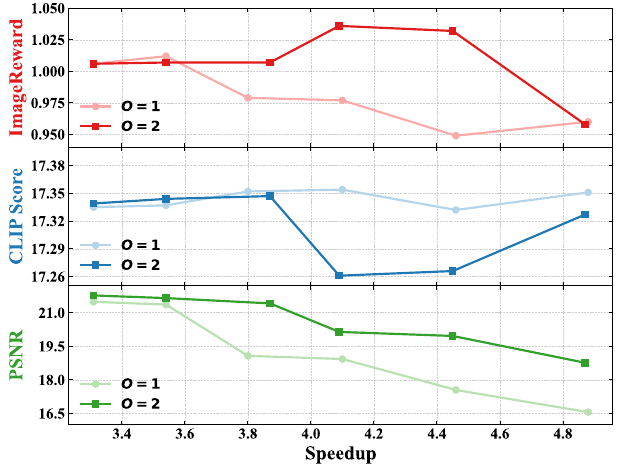}
    \caption{Performance of DPCache under varying acceleration ratios and prediction orders.}
    \label{fig:abl_k_o}
    \vspace{-0.3cm}
\end{figure}

\firstsentence{Hyperparameter analysis.}
As shown in Figure~\ref{fig:abl_k_o}, we vary the number of sampling steps $K$ to achieve different speedup ratios, and compare performance under prediction orders $O=1$ and $O=2$.
Unlike acceleration methods that rely on per-step thresholds or fixed sampling intervals, DPCache directly controls the total number of sampling steps $K$, enabling smoother and more predictable management of the speed–quality trade-off.
As the speedup ratio increases (\textit{i.e.}, $K$ decreases), all metrics evolve steadily without abrupt drops, reflecting the stability of the dynamically optimized schedule.
With respect to the prediction order, second-order prediction yields higher ImageReward and PSNR than first-order prediction, confirming that higher-order prediction yields more accurate feature extrapolation, which results in generated images of higher quality and closer alignment with the original denoising trajectory.
In contrast, first-order prediction achieves a slightly higher CLIP Score, indicating that strict trajectory adherence may come at the expense of alignment with the text prompt.

\section{Limitations}
\begin{figure}[t]
    \centering
    \includegraphics[width=\linewidth]{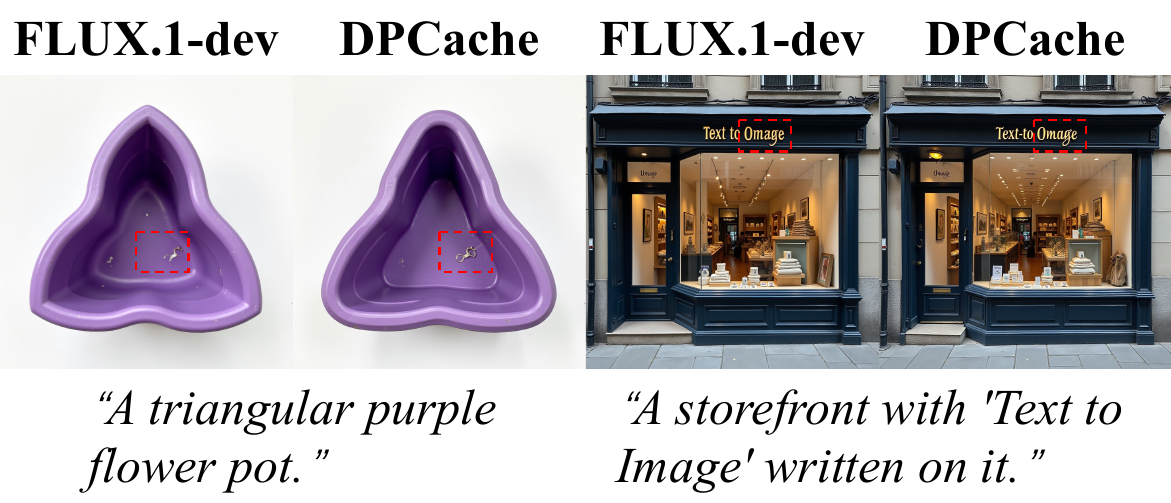}
    \caption{Failure cases of DPCache.}
    \label{fig:limitations}
\end{figure}

Although DPCache achieves significant acceleration while preserving generation quality by seeking a globally optimal sampling path, it has certain limitations. 
Since its optimization objective prioritizes fidelity to the original sampling trajectory, DPCache may sacrifice output diversity and inherit semantic or structural errors from the original model.
As shown in the left example of Figure~\ref{fig:limitations}, the original model erroneously renders “Image” as “Omage”, and DPCache retains this mistake after acceleration.
Moreover, in cases where the unaccelerated model generates plausible but imperfect results, DPCache’s emphasis on trajectory matching can inadvertently amplify minor deviations into visible artifacts.
As shown in the right example of Figure~\ref{fig:limitations}, the original model introduces spurious debris inside the flowerpot, and DPCache, by closely following the base trajectory, further amplifies this artifact. 
These observations suggest promising directions for future work. 
For instance, we could adopt input-adaptive scheduling and integrate learnable predictors to correct failures of the original model during accelerated inference.

